\documentclass{article}

\PassOptionsToPackage{numbers, sort&compress}{natbib}

\usepackage[preprint]{neurips_2021}

\usepackage[utf8]{inputenc} 
\usepackage[T1]{fontenc}    
\usepackage{hyperref}       
\usepackage{url}            
\usepackage{booktabs}       
\usepackage{amsfonts}       
\usepackage{nicefrac}       
\usepackage{microtype}      
\usepackage{xcolor}         

\usepackage[utf8]{inputenc}
\usepackage{amsmath,amsfonts,amssymb,amsthm,bbm}
\usepackage{booktabs} 
\newtheorem{theorem}{Theorem}

\usepackage{bm}
\usepackage{graphicx}
\usepackage{wrapfig}
\newcommand{\mA}{\mathbf{C}}  
\newcommand{\mB}{\mathbf{B}}
\newcommand{\mC}{\mathbf{C}}

\newcommand{\mF}{\mathbf{F}}
\newcommand{\mL}{\mathbf{L}}
\newcommand{\mQ}{\mathbf{Q}}
\newcommand{\mR}{\mathbf{R}}
\newcommand{\mW}{\mathbf{W}}

\newcommand{\mX}{\mathbf{X}}

\newcommand{\mI}{\mathbbm{I}}

\newcommand{\vx}{\mathbf{x}}
\newcommand{\vv}{\mathbf{v}}

\newcommand{\vw}{\mathbf{w}}
\newcommand{\vt}{\mathbf{t}}

\newcommand{\mU}{\mathbf{U}}
\newcommand{\mV}{\mathbf{V}}

\newcommand{\T}{\intercal}
\newcommand{\Cov}{\mathbf{C}}
\newcommand{\E}{\mathbb{E}}

\newcommand{\R}{\mathbb{R}}
\newcommand{\N}{\mathcal{N}}
\renewcommand{\O}{\mathcal{O}}

\newcommand{\tr}{\mathrm{tr}}
\renewcommand{\Cov}{\mathbf{C}}

\newcommand{\eg}{\textit{e.g.}}

\newcommand{\0}{\hspace{.5em}}
\newtheorem{lemm}{Lemma}
\newtheorem{prop}{Proposition}
\newtheorem{coro}{Corollary}

\DeclareMathOperator*{\argmin}{arg\,min}

\newcommand{\eqn}{Equation}

\newcommand{\lmm}{Lemma}
\newcommand{\thm}{Theorem}
\newcommand{\cor}{Corollary}
\newcommand{\Prop}{Proposition}  
\newcommand{\fig}{Figure}
\newcommand{\tbl}{Table}
\newcommand{\sect}{Section}
\newcommand{\appx}{Appendix}

\newcommand{\paragrapht}[1]{\noindent\textbf{#1}}  
\usepackage{algorithm}
\usepackage{algorithmic}
\usepackage{caption}
\usepackage{diagbox}
\usepackage{multirow}

\hypersetup{colorlinks,
citecolor=blue,
}

\def\beingshort
\usepackage[titletoc]{appendix}

\title{Semi-orthogonal Embedding for Efficient Unsupervised Anomaly Segmentation}

\author{%
  Jin-Hwa Kim\thanks{Corresponding author.} \\
  SK Telecom \\
  Republic of Korea \\
  \texttt{jnhwkim@sk.com} \\
  \And
  Do-Hyeong Kim \\
  SK Telecom \\
  Republic of Korea \\
  \texttt{fig.kim@sk.com} \\
  \And
  Saehoon Yi \\
  SK Telecom \\
  Republic of Korea \\
  \texttt{saehoon.yi@sk.com} \\
  \And
  Taehoon Lee \\
  SK Telecom \\
  Republic of Korea \\
  \texttt{taehoonlee@sk.com} \\
}

\begin{document}

\maketitle

\begin{abstract}
  We present the efficiency of semi-orthogonal embedding for unsupervised anomaly segmentation. The multi-scale features from pre-trained CNNs are recently used for the localized Mahalanobis distances with significant performance. However, the increased feature size is problematic to scale up to the bigger CNNs, since it requires the batch-inverse of multi-dimensional covariance tensor. Here, we generalize an ad-hoc method, random feature selection, into semi-orthogonal embedding for robust approximation, cubically reducing the computational cost for the inverse of multi-dimensional covariance tensor. With the scrutiny of ablation studies, the proposed method achieves a new state-of-the-art with significant margins for the MVTec AD, KolektorSDD, KolektorSDD2, and mSTC datasets. The theoretical and empirical analyses offer insights and verification of our straightforward yet cost-effective approach.
  
\end{abstract}

\section{Introduction}
Unsupervised anomaly segmentation is to localize the anomaly regions in the test sample while only anomaly-free samples are available in training.
So, anomaly segmentation is a more challenging task than anomaly detection, which is generally referred to detection whether a given sample has anomaly.
Anomaly segmentation is to give the visual explanation for the detected anomalies and the location for manual inspection.
In real-world problems, anomaly-free samples are usually redundant; however, anomaly samples are scarce due to manufacturing and annotation costs.
For this reason, unsupervised methods are favored while it can also provide the robustness towards unknown anomaly forms.

The reconstruction error-based methods are explored for autoencoders~\cite{bergmann2018improving,abati2019latent} and GANs~\cite{Schlegl2017}. The common idea is to train generative networks to minimize reconstruction errors learning low-dimensional features, and expect the higher error for the anomalies not presented in training than the anomaly-free.
However, the networks with a sufficient capacity could restore even anomalies causing performance degradation, although the perceptual loss function for the generative networks~\cite{bergmann2018improving} or the knowledge distillation loss for teacher-student pairs of networks~\cite{Bergmann2020,Salehi2021} achieves a limited success. 

An inspiring advance comes from a tradition, the Mahalanobis distance~\cite{Mahalanobis1936}. It measures how many standard deviations away a given sample is from the mean of a distribution of normal samples. First, one update is to use the features extracted from the CNNs pre-trained by many natural images. Second, for unsupervised anomaly segmentation, the localized Mahalanobis distance using the feature maps outperform the comparative methods~\cite{Defard2020}, which exploits a separate covariance for each location in the feature map. This method is in accordance with the assumption of a Gaussian distribution having a single mode since the distribution for every locations of feature maps tends to be a multi-modal distribution. Third, the multi-scale features enable to detect the anomalies in the interactions among the different stages~\cite{Salehi2021,Defard2020}, along with various sizes of receptive-fields in the CNNs.

However, the precision matrices for the Mahalanobis distances are required to compute for every locations in the feature map, which formulates the batch-inverse of multi-dimensional tensor where the batch size is for every locations, $H \times W$. As an ad-hoc method, \citet{Defard2020} propose to use randomly sampled features to reduce the covariance size. Our study reveals that it incurs the rank reduction when redundant features are selected with a limited budget of covariance size.

In this paper, we generalize the random feature selection to semi-orthogonal embedding as a low-rank approximation of precision matrix for the Mahalanobis distance. The uniformly generated semi-orthogonal matrix~\cite{Mezzadri2006randortho} can avoid the singular case retaining the better performance while cubically reducing the computational cost for batch-inverse. 
We achieve new state-of-the-art results for the benchmark datasets, MVTec AD~\cite{Bergmann2019}, KolektorSDD~\cite{Tabernik2019JIM}, KolektorSDD2~\cite{Bozic2021COMIND}, and mSTC~\cite{liu2018stc} while outperforming the competitive methods using reconstruction error-based~\cite{Schlegl2017,abati2019latent,bergmann2018improving} or knowledge distillation-based~\cite{Bergmann2020,Salehi2021} methods with substantial margins. Moreover, we show that our method decoupled with the pre-trained CNNs can exploit the advances of discriminative models without a fine-tuning procedure.


\section{Mahalanobis distance with multi-scale visual features}
\subsection{Mahalanobis distance for anomaly segmentation}

Let $\mX_{i,j} \in \R^{F \times N}$ be the \{i,j\}-th feature vectors from the $H \times W$ feature map extracted from a pre-trained CNNs, where $F$ is the feature size and $N$ is the number of training samples. Without loss of generality, the mean of feature vectors is zero. For each position, the covariance matrix is $\Cov_{i,j} = \frac{1}{N}\mX_{i,j}\mX_{i,j}^{\T}$, then, the squared Mahalanobis distance for a feature vector $\vx_{i,j}$ is defined as:
\begin{align}
    d_{i,j}^2 &= \vx_{i,j}^T \Cov_{i,j}^{-1} \vx_{i,j}~~\in \R^+
\end{align}

where $d_{i,j}$ indicates the anomaly score for $\vx_{i,j}$.
The Gaussian assumption of Mahalanobis distance is blessed by the localized statistics since it has a better chance to have unimodal distributions assuming that the spatial alignment of samples is done by pre-processing. 
Additionally, the multi-scale visual features are helpful to detect various sizes of anomalies~\cite{Cohen2020,Bergmann2020,Defard2020} where the feature maps from the layers with different sizes of receptive field are considered after their spatial dimensions are matched to the largest $H \times W$ by interpolation. However, the inverse of covariance matrices requires $\O(HWF^3)$, which prohibits to efficiently compute with a large $F$ of multi-scale features.

\subsection{Low-rank approximation of precision matrix}

The feature data $\mX$ is subject to low-rank approximation due to the narrower target domain for anomaly-free images than the ImageNet dataset's. The multi-scale features from different layers may also contribute to it due to the inter-dependency among the features from the layers.
Inspired by the truncated SVD of a precision matrix, a low-rank embedding of input features with $\mW \in \R^{F \times k}$, where $F > k$, is considered as follows: 
\begin{align}
    \hat{d}^2_{i,j} = \vx^\T \mW(\mW^\T \Cov_{i,j} \mW)^{-1}\mW^\T \vx
\end{align}
where the below \thm~\ref{thm:low-rank} shows the optimal $\mW^\star$ is the eigenvectors related to the $k$-smallest eigenvalues of $\Cov_{i,j}$. Notice that 1) the computational complexity of the equation is cubically reduced to $\O(HWk^3)$ set aside the cost of SVD, although which is the concern, 2) PCA embedding would fail to minimize approximation error since it uses the $k$-largest eigenvectors~\cite{Rippel2021mah}, and 3) near-zero eigenvalues may induce substantial anomaly scores. For the last, a previous work suggests to use $\Cov + \epsilon \mI$ for the inverse to avoid a possible numerical problem~\cite{Defard2020}, what we follow.

\begin{lemm}
\label{lmm:svd}\textnormal{(Truncated SVD)}~~For any $k \in 1,...,\min(F, N)$ and $\mB \in \R^{F \times k}$, let $\mU_k \in \R^{F \times k}$ be the last $k$ columns of $\mU$, which is the eigenvectors of $\mA$, and $\Sigma_k$ be a $k \times k$ diagonal matrix containing the smallest $k$ eigenvalues of a Hermitian positive definite matrix $\mA$. Then, we have that:
\begin{align}
    \min_{\mB} \|\mA^{-1} - \mB\mB^\T\|^2 &= \|\mA^{-1} - \mU_k \Sigma_k^{-1} \mU_k^\T \|^2
\end{align}
where $\|\cdot\|$ denotes the spectral norm or Frobenius norm.
\begin{proof}
The proof is through the well-known Eckart–Young theorem~\cite{eckart1936}, noting that $\mA^{-1}$ has the same eigenvectors and the inverse of eigenvalues of $\mA$, where $\mA^{-1} = \mU\Sigma^{-1}\mU^\T$.
\end{proof}
\end{lemm}

\begin{theorem}\label{thm:low-rank}
\textnormal{(Low-rank embedding of precision matrix)}~~Let $\mW \in \R^{F \times k}$, where $k \leq \min(F, N)$, be a low-rank embedding matrix. Then, we have that:
\begin{align}
    \mU_k \in \argmin_\mW \| \mA^{-1} - \mW(\mW^\T \mA \mW)^{-1}\mW^\T \|^2.
\end{align}
\begin{proof}
Because $\mW$ is a non-square matrix, it does not have a right inverse when $F > k$. If the right inverse exists, it satisfies with any $\mW$.
Letting $\mW = \mU_k$, we have:
\begin{align}
    \mW(\mW^\T \mA \mW)^{-1}\mW^\T 
    &= \mU_k(\mU_k^\T \mU\Sigma\mU^\T \mU_k)^{-1}\mU_k^\T
    = \mU_k\Sigma_k^{-1}\mU_k^\T.
\end{align}
Letting $(\mW^\T \mA \mW)^{-1}=\mV\Lambda\mV^\T$ and $\mB = \mW\mV\Lambda^{\frac{1}{2}}$, and using \lmm~\ref{lmm:svd}, we conclude the proof.
\end{proof}
\end{theorem}

However, it still needs to compute SVD or other algorithms to find the $k$-smallest eigenvectors for the multi-dimensional covariance matrix. In the following section, we will discuss a simple yet effective solution for anomaly segmentation tasks.

\section{Semi-orthogonal embedding for low-rank approximation}
\label{sec:semi}
\paragrapht{Orthogonal invariance.} The eigenvalues of a matrix $\mA$, the essence for the precise anomaly detection, are invariant concerning left or right unitary transformations of $\mA$. Moreover, \thm~\ref{orthohonal_invariance} states the invariant property of an orthogonal matrix for the inverse of $\mA$. But, since $\mW$ is a square matrix, there is no computational advantage.
\begin{prop}
\textnormal{(Orthogonal invariance)}~
\label{orthohonal_invariance}
With a consistent notation except a random orthogonal matrix $\mW_F \in \R^{F \times F}$ having orthonormal column vectors, we have that:
\begin{align}
    \mA^{-1} = \mW_F(\mW_F^\T \mA \mW_F)^{-1}\mW_F^\T.
\end{align}
\begin{proof}
The inverse of $\mW_F$ is $\mW_F^\T$. Therefore, the right term is $\mW_F\mW_F^\T \mA^{-1} \mW_F\mW_F^\T = \mA^{-1}$.
\end{proof}
\end{prop}

\paragrapht{Semi-orthogonal.} Based on this observation of the orthogonal invariance, we propose to use uniformly distributed $k$-orthonormal vectors to embed feature vectors. These vectors consist of the unitary transformation, but having at most $k$-rank constraint. The uniformly-distributed~\cite{haar1933massbegriff} orthonormal vectors are generated from Gaussian distributed random variables $\mathbf{\Omega} \in \R^{F \times k} \sim \N(0, 1)$, while the QR decomposition gives $\mathbf{\Omega} = \mQ\mR$. Since we use the same embedding to every locations of the feature map, the cost is neglectable. We get $\mW$ following the method of Mezzadri's~\cite{Mezzadri2006randortho}:
\begin{align}
    \mW = \mQ \cdot \text{sign}\big(\text{diag}(\mR)\big)~~\in \R^{F \times k}
\end{align}
where $\text{diag}(\cdot)$ returns a diagonal matrix of a given matrix and $\text{sign}(\cdot)$ returns the sign of elements in the same shape of matrix, which corrects its distribution to be uniform. 
In linear algebra, the matrix $\mW$ is called a \textit{semi-orthogonal} matrix, where $\mW^\T\mW = \mI_k \in \R^{k \times k}$, but $\mW\mW^\T \neq \mI_F \in \R^{F \times F}$.

\begin{prop}\textnormal{(Expectation of low-rank Mahalanobis distances)}~
\label{prop:exp_lowrank}
Let $\mC = \frac{1}{N}\mX \mX^\T$, $\mX \in \R^{F \times N}$, $\mW \in \R^{F \times k}$ is a matrix having $k$-orthonormal columns, where $k \leq \min(F, N)$, and $\vx$ is a column vector of $\mX$.
Then, we have that:
\begin{align}
    \E_\vx [\vx^\T \mW(\mW^\T \mC \mW)^{-1}\mW^\T \vx] = k.
\end{align}
\end{prop}
The proof is shown in \appx~\ref{appx:theory}.
The expectation of the low-rank Mahalanobis distance is the same for any semi-orthogonal matrix $\mW$.
This property is helpful to adjust the threshold depending on $k$ using the semi-orthogonal embedding.
But, be cautious that the approximation error of the precision matrix is critical to the generalization of normal samples, which impacts on the performance of anomaly segmentation.

\paragrapht{Approximation error of the precision matrix.}
The lower bound of the approximation error of a precision matrix $\mC^{-1}$ using a semi-orthogonal matrix is related to the eigenvectors corresponding $k$-smallest eigenvalues of $\mC$ as in \thm~\ref{thm:low-rank}.
Furthermore, \thm~\ref{error_bounds} states the upper bound of error using the convexity of eigenvalues in the approximation (\lmm~\ref{lmm:weighted} in \appx) and the Cauchy interlacing theorem. Please refer to \appx~\ref{appx:theory} for the proof.

\begin{theorem}
\label{error_bounds}
\textnormal{(Error bounds of low-rank precision matrix)}.~
For \textit{$\Sigma_k$ and $\mU_k$ are the diagonal matrix having the $k$-smallest eigenvalues of $\mC$ and the corresponding eigenvectors, respectively, 
and, $\Sigma_{-k}$ and $\mU_{-k}$ have the $k$-largest eigenvalues of $\mC$ and the corresponding eigenvectors, respectively,
the error bounds of the semi-orthogonal approximation of a precision matrix are that:}
\begin{align}
    \|\mC^{-1} - \mU_k \Sigma_k^{-1} \mU_k^\T \|^2 
    \leq \|\mC^{-1} - \mW(\mW^\T \mC \mW)^{-1}\mW^\T \|^2  \leq \|\mC^{-1} - \mU_{-k} \Sigma_{-k}^{-1} \mU_{-k}^\T \|^2
\end{align}
\end{theorem}
Interestingly, the assumption of flat-eigenvalues of $\mC$ where $\mC = \alpha \mI$ gives an opportunity to assess the error of the randomized approximation.
\begin{coro}
\textnormal{(Approximation error of the flat-eigenvalues)}~
\label{approximation_error}
With a consistent notation, and the flat-eigenvalues assumption of $\mC = \alpha \mI$ where $\alpha \in \R^+$. Then, we have that:
\begin{align}
    \|\mC^{-1} - \mW(\mW^\T \mC \mW)^{-1}\mW^\T \|^2
    = \frac{1}{\alpha} \| \mI_{F-k} \|^2.
\end{align}
\begin{proof}
Since the all eigenvalues are $\alpha$, we can rewrite the equation in \thm~\ref{error_bounds} as follows:
    \begin{align}
    \frac{1}{\alpha} \| \mI_{F-k} \|^2
    \leq \|\mC^{-1} - \mW(\mW^\T \mC \mW)^{-1}\mW^\T \|^2  
    \leq \frac{1}{\alpha} \| \mI_{F-k} \|^2
\end{align}
which concludes the proof.
\end{proof}
\end{coro}
In \cor~\ref{approximation_error}, 
any semi-orthogonal matrix has the same result while the interval between the bounds of approximation error are shrinked to zero for the uniform eigenvalues.
Although, in real-world problems, we do not expect the assumption is true, we cautiously remind that the batch normalization regularizes CNNs to have intermediate outputs be independently normalized~\cite{Szegedy2015}.
To evaluate our method, we show the empirical efficacy of the proposed approximation in Section~\ref{sec:experiment}.

\paragrapht{Random feature selection is a special case of the semi-orthogonal.}
If we randomly select $k$-column vectors of an identity matrix $\mI_F$, it is a special case of semi-orthogonal matrices. This is equivalent to randomly selecting $k$ features to calculate the precision matrix for Mahalanobis distance; however, it is severely vulnerable to the redundancy in features. 
If there are only $l$ of $F$ independent features where the rank is $\min(N, l)$ instead of $\min(N, F)$, the random selection of $k$ features among $F$ features has a chance that the rank is less than $k$ for the selected redundant features.
Please see the ablation study in the Experiment section and \fig~\ref{fig:mvtecad_grid_eigenplot}.


\paragrapht{Computational complexity.} The major part of the computational cost comes from the inverse of a multi-dimensional tensor. Since the use of a semi-orthogonal matrix decreases the size of the matrix from $F \times F$ to $k \times k$, the computational cost is cubically reduced to $\O(HWk^3)$. For example, the batch-SVD (the same complexity with batch-matrix inversion) of a tensor of $64 \times 64 \times 100 \times 100$ takes $191.5$ seconds, while a tensor of $64 \times 64 \times 448 \times 448$ takes $1464.6$ seconds. In the case of  $F$ is 1,796 of Wide ResNet-50-2, the computation is infeasible due to out of memory. We measure the time elapse using PyTorch 1.5 for a NVIDIA Quadro RTX 6000 with 24GB memory.

\section{Experiment}
\label{sec:experiment}
\begin{figure}[t!]
    \begin{center}
        \includegraphics[width=\textwidth]{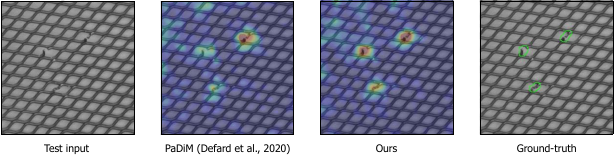}
    \end{center}
    \caption{The visualization of anomaly prediction using the jet color map after the anomaly score is clamped in [0, 10]. The green lines in the fourth image indicate the ground-truth regions. More examples can be found in \fig~\ref{fig:mvtecad_vs_visual} and \ref{fig:mvtecad_all_visual}, \appx.}
    \label{fig:mvtecad_grid_visual}
\end{figure}

\paragrapht{Datasets.} The \textit{MVTec AD} dataset~\cite{Bergmann2019}, with the CC BY-NC-SA 4.0 license, consists of five texture and ten object categories with a totally 3,629 images for training and 1,725 images for testing. 
The emerging dataset for anomaly segmentation offers the real-world categories of textures and objects having multiple types of anomalies.
The test images have single or multiple types of defects, or defect-free, while the other splits only have defect-free images. We split for validation to have 10\%, while 90\% for training. We resize the images to 256x256, evaluate on this scale, and we do not apply any data augmentation strategy being consistent with the previously published works~\cite{Bergmann2019,Bergmann2020}.
The \textit{Kolektor surface-defect dataset} (KolektorSDD)~\cite{Tabernik2019JIM} consists of the 399 images of electrical commutators, where 52 defected images are annotated for microscopic fractions or cracks on the surface of the plastic embedding in electrical commutators. The dataset is publicly available~\footnote{http://www.vicos.si/Downloads/KolektorSDD} for research and non-commercial use only.
The dataset is split by three folds, where we use only anomaly-free images for unsupervised training. 
The \textit{Kolektor surface-defect dataset 2} (KolektorSDD2)~\cite{Bozic2021COMIND} is similar with the previous one, but having more samples. The train set has 2,085 negative and 246 positive images while the test set with 894 negative and 110 positive images. For the two Kolektor datasets, we resize the images to 704x256, evaluate on this scale, and do not apply any data augmentation, for the consistent comparison in \tbl~\ref{tbl:kolektor}.
The \textit{mSTC dataset}~\cite{Venkataramanan2020} is the modified ShanghaiTech Campus (STC) dataset~\cite{liu2018stc} consisting of 13 scenes with complex light conditions and camera angles having 130 abnormal events~\footnote{\url{https://svip-lab.github.io/dataset/campus_dataset.html}}. They extract every 5-th frame of the video from each scene for training (274,515 frames) and test (42,883 frames) for unsupervised anomaly segmentation task. We randomly sample 5,000 training samples following the previous work~\cite{Defard2020}, and use the same test split. We resize the images to 256x256, evaluate on this scale, and we do not apply any data augmentation strategy being consistent with the previously published works~\cite{Venkataramanan2020,Cohen2020,Defard2020}.

\paragrapht{Metric.} The previous work~\cite{Bergmann2019} proposes a threshold-free metric based on the per-region overlap (PRO). This metric is the area under the receiver operating characteristic curve (ROC) while it takes the average of true positive rates for each connected component~\footnote{One can exploit the max pooling with a $3 \times 3$ kernel for the breadth-first search to batch-compute the markers of connected components.} in the ground truth. Because the score of a single large region can overwhelm those of small regions, the PRO promotes multiple regions' sensitivity. It calculates up to the false-positive rate of 30\% (100\% for ROC, of course).
The ROC is a natural way to cost-and-benefit analysis of anomaly decision making.

\paragrapht{Multi-scale features.} For ResNet-18, we select the layer 1, 2, and 3, having the feature sizes of 64, 128, and 256, respectively, for Wide ResNet-50-2, the feature sizes of the layer 1, 2, and 3 are 256, 512, and 1024, respectively. The feature maps from the corresponding layers are concatenated for the channel dimension after interpolating spatial dimensions to $64 \times 64$. The output map of anomaly scores using the approximated Mahalanobis distance is interpolated to $256 \times 256$ and applied the Gaussian filter with the kernel size of 4 following the previous works~\cite{Cohen2020, Defard2020}. 

\paragrapht{Ablation study 1.} In the first part of \tbl~\ref{tbl:ablation_lowrank}, the alternatives using the Mahalanobis distance are compared.
First, we confirm that the localized precision matrices, \textit{full precision (local)}, outperforms a global precision matrix, \textit{full precision (global)} for both texture and object categories with a significant margin, in spite of batch-matrix inversion. Second, the truncated SVD using the $k$-smallest eigenvalues of $\mC$ where $k \in [1, \min(F, N)]$, \textit{eigenvectors (lower)}, retains the majority of original performance, compared to its counter part using the $k$-largest eigenvalues, \textit{eigenvectors (higher)} as predicted in \thm~\ref{thm:low-rank}. These two alternatives provide the lower and upper bounds of the semi-orthogonal approximation error. Notice that $F$=1,792 for Wide ResNet-50-2, with the multi-scale consideration, is prohibitively expensive in both computation and memory for batch-matrix inversion.

\paragrapht{Ablation study 2.} The second set of experiments is on the choice of layers providing input features. For ResNet-18, the feature sizes of layer 1, 2, and 3 are 64, 128, and 256, respectively, denoted by $F_a$, $F_b$, and $F_c$ in \tbl~\ref{tbl:ablation_lowrank}. Notice that $F_b$ and $F_c$ are larger than $k=100$. The results confirm the multi-scale approach using the multiple layers is crucial in anomaly segmentation. 
Especially, the layer 2 and 3 underperform our \textit{semi-orthogonal} method with a higher computational complexity.

\paragrapht{Ablation study 3.} The third part of \tbl~\ref{tbl:ablation_lowrank} compares the alternatives with the embedding size $k$ of 100. The Gaussian random-valued embeddings, \textit{Gaussian}, significantly deteriorates the performance. 
The PaDiM~\cite{Defard2020} uses the same approximation as in \textit{sampled features}, however, our careful implementation gets a stronger baseline with 0.912 compared with their 0.905. One of reasons is the pre-processing where we do not use the center cropping since the center area does not perfectly cover the anomalies in some cases unlike their assumption. Notice that our evaluation protocol is consistent with previous works. Surprisingly, the approximation using our semi-orthogonal matrix outperforms the comparative methods with 0.924 considerably retaining the performance of \textit{full precision (local)} (0.934) with 1\% of computation and 5\% of memory complexities of those. For Wide ResNet-50-2, the ratio is more conversing to zero.
Also, we verify that the average of the standard deviations per category with five random semi-orthogonal matrices is lower than that of \textit{sampled features}.

\begin{table*}
  \centering
  \caption{Ablation study on low-rank methods for the anomaly segmentation task of the MVTec AD using the per-region-overlap (PRO). We use ResNet-18 to extract features and the target rank $k$ is 100 where $F_a < k < F_b < F_c < F$. The computational complexity is cubically dependent on $F$, $F_{a-c}$, or $k$. \textit{Std.} indicates the average of the standard deviations per category with five random seeds. 
} 
  \label{tbl:ablation_lowrank}
  \begin{tabular}{lccccc}
  \toprule
  Model                             & Complexity & Texture   & Object    & Overall   & Std. \\ \midrule
  Full-rank (global)           & $\O(F^3)$  & .866      & .899      & .888      & -    \\ 
  Full-rank (local)            & $\O(HWF^3)$& .920      & \bf{.941} & \bf{.934} & -    \\ 
  Eigenvectors (higher)             & $\O(HWF^3)$& .886      & .896      & .893      & -    \\
  Eigenvectors (lower)              & $\O(HWF^3)$& \bf{.921} & .939      & .933      & -    \\ \midrule

  Layer 1                           &$\O(HWF_a^3)$& .880      & .901      & .894      & -    \\ 
  Layer 2                           &$\O(HWF_b^3)$& \bf{.903} & \bf{.921} & \bf{.915} & -    \\ 
  Layer 3                           &$\O(HWF_c^3)$& .883      & .916      & .905      & -    \\ \midrule

  Gaussian                          & $\O(HWk^3)$ & \bf{.915} & .872      & .886      & .003      \\
  Sampled features~\cite{Defard2020}& $\O(HWk^3)$ & .888      & .924      & .912      & .009      \\
  Semi-orthogonal (ours)            & $\O(HWk^3)$ & .909      & \bf{.931} & \bf{.924} & \bf{.002} \\
  \bottomrule
  \end{tabular}
\end{table*}

\begin{table*}
  \centering
  \caption{Comparison with the state-of-the-art for the anomaly segmentation task of the MVTec AD dataset using the two metrics, PRO and ROC. Please see the text for details.}
  \label{tbl:mvtecad_pro}
  \begin{tabular}{lccccc}
  \toprule
        &          & \multicolumn{3}{c}{PRO} & ROC   \\
  \cmidrule(lr){3-5} \cmidrule(lr){6-6}
  Model & Backbone & Texture & Object & Overall & Overall \\ \midrule
  Autoencoder (SSIM)~\cite{Bergmann2019} & - & .567 & .758 & .694 & .870 \\
  Autoencoder (L2)~\cite{Bergmann2019}   & - & .696 & .838 & .790 & .820 \\
  VAE~\cite{Defard2020}                  & - & .499 & .714 & .642 & .744  \\
  AnoGAN~\cite{Schlegl2017}              & - & .274 & .533 & .443 & .743  \\ 
  CAVGA~\cite{Venkataramanan2020}        & DC-GAN~\cite{radford2016dcgan} & - & - & - & .85\0  \\ \midrule
  Multi-KD~\cite{Salehi2021}             & VGG-16           & -    & -    & -    & .907 \\
  FCDD~\cite{Liznerski2021}              & -                & -    & -    & -    & .92\0 \\
  Patch-SVDD~\cite{Yi2020}               & -                & -    & -    & -    & .957 \\
  Uninformed Student~\cite{Bergmann2020} & ResNet-18        & .794 & .889 & .857 & -    \\
  SPADE~\cite{Cohen2020}                 & Wide ResNet-50-2 & .884 & .934 & .917 & .965 \\
  PaDiM (k=100)~\cite{Defard2020}          & ResNet-18        & .913 & .894 & .901 & .967 \\
  PaDiM (k=550)~\cite{Defard2020}          & Wide ResNet-50-2 & .932 & .916 & .921 & .975 \\ \midrule
  Ours (k=100) & MobileNetV3-Small& .924      & .885      & .898      & .968 \\
  Ours (k=100) & MobileNetV3-Large& .923      & .899      & .909      & .972 \\
  Ours (k=100) & ResNet-18        & .909      & .931      & .924      & .975 \\
  Ours (k=100) & Wide ResNet-50-2 & .925      & .938      & .934      & .979 \\
  Ours (k=300) & Wide ResNet-50-2 & \bf{.934} & \bf{.946} & \bf{.942} & \bf{.982} \\
  \bottomrule
  \end{tabular}
\end{table*}

\begin{table*}[t!]
  \centering
  \caption{The ROC results for the unsupervised anomaly segmentation task using the KolektorSDD and KolektorSDD2 datasets. We report the score for each fold of the KolektorSDD dataset, their mean and standard deviation (Std.), and the mean score for the KolektorSDD2 dataset with the standard deviation with three random seeds.
  Notice that we reproduce the scores of PaDiM~\cite{Defard2020} with the same setting with ours except the method of approximation. We use ResNet-18 with k=100 for the setting.
  } 
  \label{tbl:kolektor}
  \begin{tabular}{lccccc}
  \toprule
  Model                                  & Fold 1    & Fold 2    & Fold 3    & Mean $\pm$ Std.      & KolektorSDD2         \\ \midrule
  Uninformed student~\cite{Bergmann2020} & .904      & .883      & .902      & .896 $\pm$ .012      & .950 $\pm$ .005      \\
  PaDiM~\cite{Defard2020}                & .939      & .935      & .962      & .945 $\pm$ .015      & .956 $\pm$ .000      \\
  Semi-orthogonal (ours)                 & \bf{.953} & \bf{.951} & \bf{.976} & \bf{.960 $\pm$ .014} & \bf{.981 $\pm$ .000} \\
  \bottomrule
  \end{tabular}
\end{table*}

\paragrapht{State-of-the-art of MVTec AD.} We achieve a new state-of-the-art for the MVTec AD in the two major metrics, PRO and ROC, in the comparison with competing methods, with significant margins. The results consistently show that Mahalanobis distance-based methods~\cite{Defard2020} outperform the reconstruction error-based~\cite{bergmann2018improving,Bergmann2019,Schlegl2017} and knowledge distillation-based methods~\cite{Bergmann2020}.
Notably, using Wide ResNet-50-2, our method significantly outperforms PaDiM with less computational and memory complexities with $k$=300.
It suggests that our method can readily exploit more powerful backbone networks while the computational cost is efficiently controlled by $k$, without any fine-tuning of backbone networks.
The scores of the total 15 categories can be referred in \tbl~\ref{tbl:mvtecad_pro_category}, \appx.

\paragrapht{State-of-the-art of KolektorSDD.} In \tbl~\ref{tbl:kolektor}, we compare our method with the other comparative methods using the KolektorSDD and KolektorSDD2 datatsets. Although the third fold of the KolektorSDD dataset tends to have slightly higher score than others which impacts on the standard deviation, our method consistently outperforms the others across all folds. 
Using the more samples in the KolektorSDD2, the performance of the uninformed student~\cite{Bergmann2020} is notably improved; however, the localized Mahalanobis-based methods outperform it, and our method shows the consistence.
For this comparison, we reproduce the uninformed student~\cite{Bergmann2020} and PaDiM~\cite{Defard2020}. Notice that we use the same setting of ResNet-18 with k=100 for the PaDiM except the approximation method. For the uninformed student~\cite{Bergmann2020}, we use their model with the receptive size of 33 $\times$ 33 for the best result. We follow the other training settings.

\begin{table*}[t!]
  \centering
  \caption{The ROC results for the unsupervised anomaly segmentation task using the mSTC dataset. 
  Our method use ResNet-18 with k=100 for the fair comparison.
  } 
  \label{tbl:stc}
  \begin{tabular}{lcccc}
  \toprule
  Model     & CAVGA-RU~\cite{Venkataramanan2020} & SPADE~\cite{Cohen2020} & PaDiM~\cite{Defard2020} & Ours \\ \midrule
  ROC     & .85\0    & .899    & .912    & \bf{.921} \\
  \bottomrule
  \end{tabular}
\end{table*}

\paragrapht{State-of-the-art of mSTC.} In \tbl~\ref{tbl:stc}, our method consistently outperforms the comparative methods in the unsupervised abnormal event segmentation task, while achieving a new state-of-the-art. Notice that this dataset has different domain from the other datasets since the manufacturing product images tend to have a similar shape and are center-aligned.

\paragrapht{Visualization.} \fig~\ref{fig:mvtecad_grid_visual} visualizes an example from the \textit{Grid} category. In this case, our method shows a better prediction than the previous state-of-the-art~\cite{Defard2020}, detecting three small regions of anomalies. Please remind that the metric of PRO emphasizes to detect all small regions in the ground-truth. More examples can be found in \fig~\ref{fig:mvtecad_all_visual}, \appx.

\section{Discussion}
\label{sec:discussion}

\begin{figure}
\begin{minipage}{.32\textwidth}
    \begin{center}
        \includegraphics[width=\textwidth]{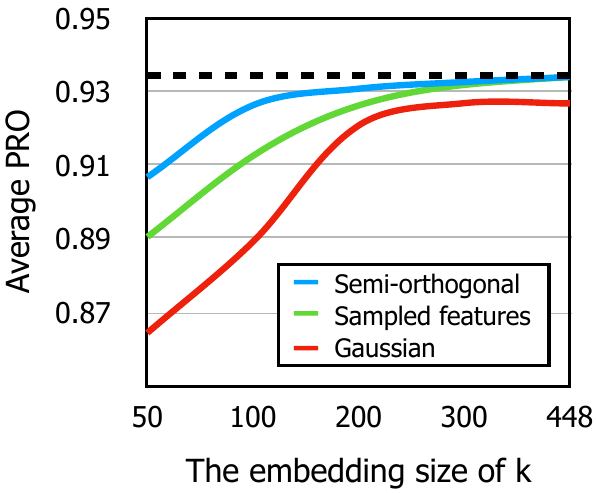}
    \end{center}
    \caption{The average PRO with respect to the embedding size $k$.}
    \label{fig:mvtecad_rank}
\end{minipage}
\hspace{1em}
\begin{minipage}{.64\textwidth}
    \begin{center}
        \includegraphics[width=\textwidth]{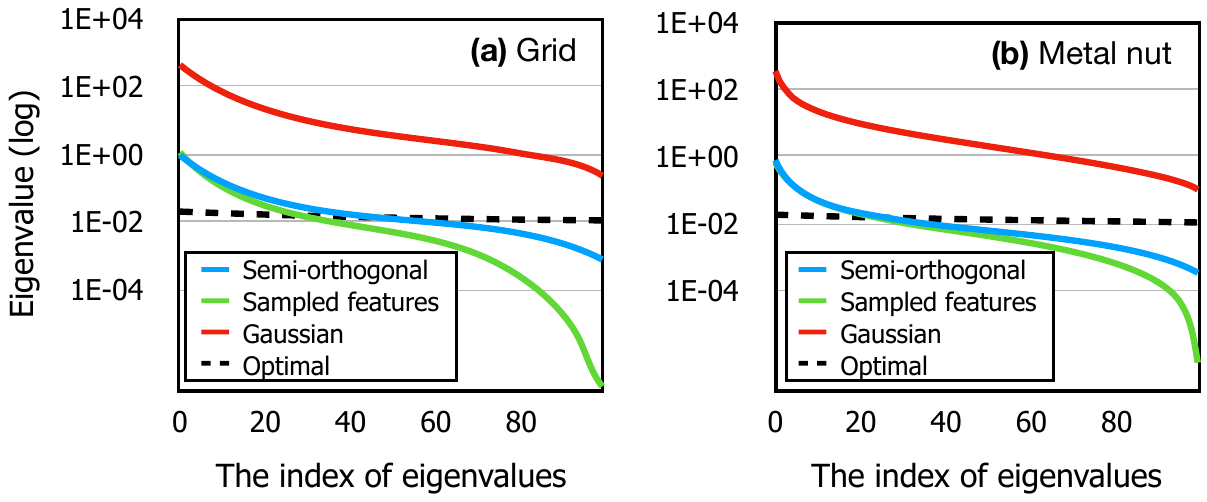}
    \end{center}
    \caption{The eigenvalues of the comparative embedded covariance matrices $\mW^\T\mC\mW$ of the \textit{Grid} and \textit{Metal nut} categories.}
    \label{fig:mvtecad_grid_eigenplot}
    \end{minipage}
\end{figure}

\paragrapht{Parsimonious of low-rank approximations.}
The average PRO scores from the fifteen cateogries of MVTec AD with respect to the embedding size of $k$ using the backbone of ResNet-18 are shown in \fig~\ref{fig:mvtecad_rank}.
When the $k$ is smaller than 200 the performance gaps are distinctive among comparative methods. Among them, the approximation with our \textit{semi-orthogonal} embedding achieves the best performance with a significant margin when the $k$ is less than 100.
The worst method is the \textit{Gaussian} which fails to retain the optimal performance even with the full embedding size of $F$.
The dashed line denotes the best achievable score with the full precision matrix $\mC^{-1}$.

\paragrapht{Rank collapse in feature selection.}
In \tbl~\ref{tbl:ablation_lowrank}, the \textit{sampled features}~\cite{Defard2020} 
is a special case of ours where the semi-orthogonal matrix $\mW$ is randomly selected $k$ columns of the $F \times F$ identity matrix.
This strategy has a drawback when input features have some degree of redundancy, which is prone to have the rank lower than $k$ for the chance of redundant selection.
For which, \fig~\ref{fig:mvtecad_grid_eigenplot} shows the expectation of eigenvalues with respect to the location of a feature map for \textit{Grid} and \textit{Metal nut}.
The plots of \textit{sampled features} consistently show rank collapses starting from the around 80 to 90-th eigenvalues (for $k$ is 100) compared with those of \textit{semi-orthogonal}, which is the empirical evidence indicating the vulnerability of feature sampling strategy.
The \textit{Grid} and \textit{Metal nut} categories have only 81.13\% ($\pm0.71$) and 79.34\% ($\pm1.85$) of all eigenvalues (which are not expected values) higher than 1e-4, respectively. Whereas, \textit{Gaussian} and \textit{semi-orthogonal} have 100\% of eigenvalues (full-rank) higher than 1e-4.
Although \textit{Gaussian} does not induce the rank collapse, the eigenvalues are randomly scaled deviating from the original distribution of eigenvalues, which may interfere the detection of anomalies.
Note that the dashed line indicates the plot of the expected $k$-smallest eigenvalues with respect to the location of a feature map as a reference.

\paragrapht{In defense of using the localized Mahalanobis distances.}
Compared to the other approaches, the Mahalanobis distance considers the second moment (covariance) of feature statistics to measure the degree of anomaly. The assumption of Gaussian distribution is backed by the localized statistics in the feature map extracted by pre-trained CNNs. The computational cost of inverse for the multi-dimensional precision tensor is cubically reduced by the proposed semi-orthogonal embedding successfully retaining the empirical performances as in \tbl~\ref{tbl:ablation_lowrank} and \ref{tbl:mvtecad_pro}.
The decoupled with the backbone networks is the important advantage for small datasets that does not require the fine-tuning of feature extractor.
Moreover, it can be shown that the knowledge distillation loss is related to the Mahalanobis distance if linear models are used (\appx~\ref{appx:kd}). This speculation weights on the future work to advance efficient Mahalanobis distance-based methods for anomaly segmentation tasks.

\paragrapht{Limitation.}
The localized Mahalanobis distance for anomaly segmentation requires the batch-inverse of multi-dimensional tensor. Although it significantly outperforms the global Mahalanobis distance in the ablation study 1 (\sect~\ref{sec:experiment}) for both texture and object categories, the computational cost may increase depending on the size of a feature map, \eg, 4K high-resolution images.

\paragrapht{Social impact.}
The automation may induce the reduction of workers, although it can improve the quality assurance of manufacturing and relieve the workers from repetitive labors.

\section{Related work}
With growing attention to anomaly detection and segmentation tasks, comprehensive reviews on this topic, including recent deep learning approaches, are available~\cite{pimentel2014review,Pang2020review,Perera2021review}. 
Notice that we focus on unsupervised anomaly segmentation, locating any anomaly region in a test image where only anomaly-free images are available while training.
Therefore, in this section, we would like to highlight the following related works cohesively.

\paragrapht{Reconstruction errors.} Generative models such as autoencoders and GANs are employed for the task~\cite{Schlegl2017, bergmann2018improving, abati2019latent}. These methods are based on an optimistic view on reconstruction errors that anomalies cause higher reconstruction errors than the others. Notice that a model can generate even anomalies for its robust reconstruction capability. A perceptual loss function~\cite{bergmann2018improving} proposed to overcome this limitation; however, it requires domain knowledge to design the function and has shown a little improvement compared with the other recent approaches~\cite{burlina2019s,Bergmann2020}.

\paragrapht{Embedding feature similarity.} One of the breakthroughs comes from the utilization of pre-trained CNNs. An early work~\cite{andrews2016transfer} uses the VGG networks~\cite{Simonyan2015vgg} for anomaly image detection tasks.
Especially, a work~\cite{burlina2019s} explicitly shows that learned discriminative embeddings are better than generative models.
One of successful approaches~\cite{Bergmann2020,Salehi2021} is to use the knowledge distillation~\cite{Hinton2014}.
This method utilizes the knowledge distillation loss instead of reconstruction errors assuming that the knowledge distillation loss would increase when anomalies appear.
However, with a sufficient network capacity, the loss would vanish when the student networks behavior similarly with the teacher networks. They made pre-designed small CNNs for the teacher networks, distilled from the pre-trained ResNet-18 networks.

\paragrapht{Mahalanobis distance.}
Mahalanobis distance is a metric that measures the distance between two points discounted by a covariance.
Notably, PaDiM~\cite{Defard2020}, the current state-of-the-art anomaly segmentation method for the MVTec AD dataset~\cite{Bergmann2019}, utilizes 1) the discriminative features from pre-trained CNNs, 2) multi-resolution features, similarly to SPADE~\cite{Cohen2020}, for robust detection, and 3) uses separate statistics per feature location for the unimodal Gaussian assumption of Mahalanobis distance.
Since the separate statistics require the inverse of multi-dimensional covariance tensor, the key problem in this approach was the approximation of the inverse of covariance tensor.
They proposed the random selection of features, while we argue the shortcoming of rank collapse and propose a better solution to use a uniformly generated semi-orthogonal matrix.

\paragrapht{Orthogonal embedding.}
The Johnson-Lindenstrauss lemma~\cite{johnson1984} examines the embeddings from high-dimensional into low-dimensional Euclidean space, in a way that the distances among the samples are virtually preserved, \eg, orthogonal projection. A line of works uses random orthogonal matrices to approximate a Gram matrix in the kernel methods~\cite{Choromanski2017} or proposes the orthogonal low-rank embedding loss to reduce intra-class variance and enforce inter-class margin simultaneously for classification tasks~\cite{Lezama2018}. Related to anomaly detection, the LRaSMD~\cite{zhang2016} is proposed to solve a hyperspectral anomaly detection problem using a low-rank and sparse matrix decomposition of data. However, it does not consider a large multi-dimensional covariance matrix, which needs to approximate the Mahalanobis distance for anomaly segmentation.

\section{Conclusion}
We propose the semi-orthogonal embedding method for the low-rank approximation of the localized Mahalanobis distance reducing the computational cost for the batch-inverse of covariance matrices without batch-SVD computation. 
We show that the proposed method is the generalization of the random feature selection method in the previous work~\cite{Defard2020}, while retaining the better performance by avoiding redundant sampling. 
We achieve new state-of-the-arts for the benchmark dataset, MVTec AD~\cite{Bergmann2019}, KolektorSDD~\cite{Tabernik2019JIM}, KolektorSDD2~\cite{Bozic2021COMIND}, and mSTC~\cite{liu2018stc}, outperforming the competitive methods using reconstruction error-based~\cite{Schlegl2017,abati2019latent,bergmann2018improving} or knowledge distillation-based methods~\cite{Bergmann2020,Salehi2021} with significant margins. 
We emphasize that our method 1) implicitly considers multi-scale receptive fields exploiting the feature maps from multiple layers of CNNs, 2) uses the localized Mahalanobis distance for fine-grained anomaly segmentation via an interpretable metric, and 3) is decoupled with the pre-trained CNNs that can exploit the advances of discriminative models without fine-tuning.


\bibliography{main}
\bibliographystyle{unsrtnat}

\newpage

\setcounter{section}{0}
\renewcommand\thesection{\Alph{section}}
\renewcommand\thesubsection{\thesection.\arabic{subsection}}

\section{Theoretical analysis}
\label{appx:theory}
{\bf\Prop~\ref{prop:exp_lowrank}. restated. }\textnormal{(Expectation of low-rank Mahalanobis distances)}.~
\textit{Let $\mC = \frac{1}{N}\mX \mX^\T$, $\mX \in \R^{F \times N}$, $\mW \in \R^{F \times k}$ is a matrix having $k$-orthonormal columns, where $k \leq \min(F, N)$, and $\vx$ is a column vector of $\mX$.
Then, we have that:}
\begin{align}
    \E_\vx [\vx^\T \mW(\mW^\T \mC \mW)^{-1}\mW^\T \vx] = k.
\end{align}
\begin{proof}
First, we simplify the equation using $\mF=\mX^\T\mW$:
\begin{align}
    \E_\vx [\vx^\T \mW(\mW^\T \mC \mW)^{-1}\mW^\T \vx] 
    &= \frac{1}{N}\text{tr}\Big( \mX^\T \mW \big( \mW^\T (\frac{1}{N}\mX\mX^\T) \mW \big)^{-1}\mW^\T \mX \Big) \nonumber \\
    &= \text{tr}\big( \mX^\T \mW ( \mW^\T \mX\mX^\T \mW )^{-1}\mW^\T \mX \big) \nonumber \\
    &= \text{tr}( \mF \mF^\dagger )
\end{align}
where $\mF^\dagger$ is the left Moore–Penrose inverse of $\mF$ where $\mF\mF^\dagger \neq \mI$. Using the singular value decomposition of $\mF = \mU\Sigma\mV^\T$ and the invariant of trace, 
\begin{align}
    \text{tr}( \mF \mF^\dagger )
    = \text{tr} ( \mU\Sigma\mV^\T \mV\Sigma^\dagger\mU^\T ) 
    = \text{tr} ( \Sigma\Sigma^\dagger )
    = k.
\end{align}
where $\Sigma$ has $k$ non-zero elements. 

Notice that $\mW$ consists of any $k$-eigenvectors of $\mC$, which are orthonormal, holds the same result.
\end{proof}

\begin{lemm}
\label{lmm:weighted}\textnormal{(Convex combination of eigenvalues)}~~For the semi-orthogonal matrix $\mW \in \R^{F \times k}$, any eigenvalue of $\mW^\T\mC\mW$ is the convex combination of the eigenvalues of $\mC$:
\begin{align}
    \hat{\lambda} = \hat{\vv}^\T \mW^\T\mC\mW \hat{\vv} = \sum_i a_i \lambda_i
\end{align}
where $\hat{\lambda}$ is an eigenvalue of $\mW^\T\mC\mW$, $\hat{\vv}$ is the corresponding unit eigenvector to $\hat{\lambda}$, $\{\lambda_i\}$ are the eigenvalues of $\mC$, $\sum_i a_i=1$, and $a_i \geq 0$.
\begin{proof}
The eigenvalue decomposition of $\mC$ yields $\mU \Sigma \mU^\T$ as follows:
\begin{align}
    \hat{\vv}^\T \mW^\T\mC\mW \hat{\vv}
    &= \hat{\vv}^\T \mW^\T\mU \Sigma \mU^\T\mW \hat{\vv} \\
    &= \sum_i (\hat{\vv}^\T \mW^\T\mU \circ \hat{\vv}^\T \mW^\T\mU)_i \lambda_i
\end{align}
while the weight for $\lambda_i$ is non-negative by square, and the sum of weight is one as follows:
\begin{align}
    \sum_i (\hat{\vv}^\T \mW^\T\mU \circ \hat{\vv}^\T \mW^\T\mU)_i 
    = \hat{\vv}^\T \mW^\T\mU \mU^\T\mW \hat{\vv}
    = 1
\end{align}
where $\mU\mU^\T=\mI_F$ and $\mW^\T\mW=\mI_k$, which concludes the proof.
\end{proof}
\end{lemm}

{\bf\thm~\ref{error_bounds}. restated. }\textnormal{(Error bounds of low-rank precision matrix)}.~
For \textit{$\Sigma_k$ and $\mU_k$ are the diagonal matrix having the $k$-smallest eigenvalues of $\mC$ and the corresponding eigenvectors, respectively, 
and, $\Sigma_{-k}$ and $\mU_{-k}$ have the $k$-largest eigenvalues of $\mC$ and the corresponding eigenvectors, respectively,
the error bounds of the semi-orthogonal approximation of a precision matrix are that:}
\begin{align}
    \|\mC^{-1} - \mU_k \Sigma_k^{-1} \mU_k^\T \|^2 
    \leq \|\mC^{-1} - \mW(\mW^\T \mC \mW)^{-1}\mW^\T \|^2  \leq \|\mC^{-1} - \mU_{-k} \Sigma_{-k}^{-1} \mU_{-k}^\T \|^2
\end{align}
\begin{proof}
The lower bound comes from \thm~\ref{thm:low-rank}. 

For the upper bound, we start with the case of the Frobenius norm, 
\begin{align}
    \|\mC^{-1} - \mW(\mW^\T \mC \mW)^{-1}\mW^\T \|^2
    & = \|\mC^{-1} \|^2 + \| \mW(\mW^\T \mC \mW)^{-1}\mW^\T \|^2 \nonumber\\
    &    - 2 \tr(\mC^{-1}\mW(\mW^\T \mC \mW)^{-1}\mW^\T)
\end{align}
where the last term is the Frobenius inner product. Now, we concern about the function of $\mW$ as follows:
\begin{align}
    \label{eqn:target_trace}
    &\| \mW(\mW^\T \mC \mW)^{-1}\mW^\T \|^2 - 2\tr(\mC^{-1}\mW(\mW^\T \mC \mW)^{-1}\mW^\T) \nonumber \\
    &~~~~= \| (\mW^\T \mC \mW)^{-1} \|^2 - 2\tr(\mC^{-1}\mW(\mW^\T \mC \mW)^{-1}\mW^\T)~~~~~~~~~~\text{(cyclic invariance)} \nonumber \\
    &~~~~= \| (\mW^\T \mC \mW)^{-1} \|^2 - 2\tr(\mW^\T\mC^{-1}\mW(\mW^\T \mC \mW)^{-1})~~~~~~~~~~\text{(cyclic invariance)} \nonumber \\
    &~~~~= \tr\big((\mW^\T \mC \mW)^{-2}\big) - 2\tr\big(\mW^\T\mC^{-1}\mW(\mW^\T \mC \mW)^{-1}\big).~~~~~~~\text{(by definition)}
\end{align}
By the way, letting $(\mW^\T\mC\mW)^{-1} = \hat{\mV}\hat{\mL}\hat{\mV}^\T$, the last trace can be rewritten as:
\begin{align}
    &\tr\big(\mW^\T\mC^{-1}\mW(\mW^\T \mC \mW)^{-1}\big) \nonumber\\
    &~~~~= \tr\big(\mW^\T\mC^{-1}\mW \hat{\mV}\hat{\mL}\hat{\mV}^\T\big) \nonumber\\
    &~~~~= \tr\big(\hat{\mV}^\T\mW^\T\mC^{-1}\mW \hat{\mV}\hat{\mL}\big) ~~~~~~~~~~~~~~~~~~~\text{(cyclic invariance)} \nonumber \\
    &~~~~= \tr\big(\hat{\mV}^\T\mW^\T (\mU\Sigma^{-1}\mU^\T) \mW \hat{\mV}\hat{\mL}\big) ~~~~~~~\text{(by definition)}
\end{align}
where, using \lmm~\ref{lmm:weighted}, the eigenvalue of $(\mW^\T\mC\mW)^{-1}$ in $\hat{\mL}$ is the harmonic weighted sum of the eigenvalues of $\mC^{-1}$. Moreover, the diagonal elements of $\hat{\mV}^\T\mW^\T (\mU\Sigma^{-1}\mU^\T) \mW \hat{\mV}$ is the arithmetic weighted sum of the eigenvalues of $\mC^{-1}$ as follows:
\begin{align}
    &\big( \hat{\mV}^\T\mW^\T (\mU\Sigma^{-1}\mU^\T) \mW \hat{\mV} \big)_{ii} \nonumber \\
    &~~~~= \hat{\vv}_i^\T\mW^\T (\mU\Sigma^{-1}\mU^\T) \mW \hat{\vv}_i \nonumber \\
    &~~~~= \sum_j (\hat{\vv}_i^\T \mW^\T\mU \circ \hat{\vv}_i^\T \mW^\T\mU)_i \lambda_i^{-1}
\end{align}
where $\{\hat{\vv}_i\}$ are the eigenvectors of $(\mW^\T\mC\mW)^{-1}$ and $\{\lambda_i\}$ are the eigenvalues of $\mC$. Note that the weights for the harmonic weighted sum and the arithmetic weighted sum are the same.
Therefore, we use the weighted arithmetic-harmonic inequality~\cite{maze2009note} as follows:
\begin{align}
    \tr\big(\hat{\mV}^\T\mW^\T (\mU\Sigma^{-1}\mU^\T) \mW \hat{\mV}\hat{\mL}\big) 
    &\ge \tr(\hat{\mL}\hat{\mL}) \nonumber \\
    &= \tr( \hat{\mV}\hat{\mL}\hat{\mV}^\T \hat{\mV}\hat{\mL}\hat{\mV}^\T ) \nonumber \\
    &= \tr\big( (\mW^\T\mC\mW)^{-1} (\mW^\T\mC\mW)^{-1} \big)
\end{align}
Now, we use $\tr\big((\mW^\T\mC\mW)^{-2}\big) \leq \tr\big(\mW^\T\mC^{-1}\mW (\mW^\T\mC\mW)^{-1}\big)$ to simplify as follows:
\begin{align}
    &\tr\big((\mW^\T \mC \mW)^{-2}\big) - 2\tr\big(\mW^\T\mC^{-1}\mW(\mW^\T \mC \mW)^{-1}\big) \nonumber\\
    &~~~~\leq \tr\big((\mW^\T \mC \mW)^{-2}\big) - 2~\tr\big((\mW^\T \mC \mW)^{-2}\big) \nonumber \\
    &~~~~= -\tr\big((\mW^\T \mC \mW)^{-2}\big).
\end{align}
The Cauchy interlacing theorem states that the $i$-th eigenvalue of $\mW^\T\mC\mW$ is less than or equal to $i$-th eigenvalue of $\mC$ and greater than or equal to the $F-k+i$-th eigenvalue of $\mC$, where the eigenvalues are in descending order. For the upper bound, $\mW$ is chosen for the corresponding eigenvectors to the $k$-largest eigenvalues of $\mC$. Therefore, the upper bound is as follows:
\begin{align}
    &\|\mC^{-1} \|^2 -\tr\big((\mU_{-k}^\T \mC \mU_{-k})^{-2}\big) \nonumber \\
    &~~~~= \tr(\mC^{-2}) - \tr\big((\mU_{-k}^\T \mC \mU_{-k})^{-2}\big) \nonumber \\
    &~~~~= \tr\big(\mC^{-2} - \mU_{-k} \Sigma_{-k}^{-2} \mU_{-k}^\T \big) \nonumber \\
    &~~~~= \tr\big(\mC^{-2} - 2\mC^{-1}\mU_{-k} \Sigma_{-k}^{-1} \mU_{-k}^\T + \mU_{-k} \Sigma_{-k}^{-2} \mU_{-k}^\T \big) \nonumber \\
    &~~~~= \tr\big((\mC^{-1} - \mU_{-k} \Sigma_{-k}^{-1} \mU_{-k}^\T)^2 \big) \nonumber \\
    &~~~~= \|\mC^{-1} - \mU_{-k} \Sigma_{-k}^{-1} \mU_{-k}^\T \|^2.
\end{align}

In the case of the spectral norm, 
\begin{align}
    &\|\mC^{-1} - \mW(\mW^\T \mC \mW)^{-1}\mW^\T \|^2 \nonumber \\
    &~~~~=\big( \max_{\|\vv\|=1} \vv^\T \mC^{-1} \vv - \vv^\T \mW(\mW^\T \mC \mW)^{-1}\mW^\T \vv \big)^2 \nonumber \\
    &~~~~\leq \big( \max_{\|\vv\|=1} \vv^\T \mC^{-1} \vv \big)^2
\end{align}
where the inequality comes from the Cauchy interlacing theorem. The difference in the max function is greater than or equal to zero since there exists at least one instance that, if we choose $\vv$ for the spectral norm of $\mC^{-1}$, $\vv^\T \mW(\mW^\T \mC \mW)^{-1}\mW^\T \vv$ is less than or equal to the spectral norm of $\mC^{-1}$. The approximation has non-negative eigenvalues. Therefore, the upper bound is the square of the smallest eigenvalue of $\mC$. Remind that the smallest eigenvalue of $\mC$ is the largest eigenvalue of $\mC^{-1}$. 

Now, we conclude the proof.
\end{proof}

\section{Knowledge distillation to Mahalanobis distance}
\label{appx:kd}
The uninformed students~\cite{Bergmann2020} use the knowledge distillation loss~\cite{Hinton2014} to detect anomaly regions. We speculate that the knowledge distillation loss, when early stopping is used, connects to the Mahalanobis distance in \textit{linear models}.

Let the training data $\E[\vx]=0$, the teacher model is $\vw^\star$, the student model is $\vw_n$ at the $n$-iteration of learning. The mean squared error is define as:
\begin{align}
    \mathcal{L}^{(n)} = \E_{\vx \sim \mathcal{D}} \big[ 
            \frac{1}{2} \| \vw^{\star\T}\vx - \vw_n^\T\vx \|^2  
        \big].
\end{align}
Then, the gradient with respect to the parameter of the student is as follows:
\begin{align}
    \frac{\partial \mathcal{L}^{(0)}}{\partial \vw_0} 
    &= \E \big[ \vx(\vw_0^\T\vx - \vw^{\star\T}\vx)^\T \big] \nonumber \\
    &= \E[ \vx\vx^\T] (\vw_0 - \vw^\star) \nonumber \\
    &= \mC (\vw_0 - \vw^\star)
\end{align}
where the SGD with the learning rate $\eta$ updates the parameter as follows:
\begin{align}
    \vw_1 \leftarrow \vw_0 - \eta \mC (\vw_0 - \vw^\star)
\end{align}
The analytical solution of the student parameter at the $n$-step is available via the Neumann series, which is the geometric series for matrices.
\begin{align}
    \vw_n = \big( \mI - (\mI - \eta\mC)^n \big) \vw^\star + (\mI - \eta\mC)^n \vw_0
\end{align}
Then, the loss is rewritten as:
\begin{align}
    \mathcal{L}^{(n)} = \E_{\vx \sim \mathcal{D}} \big[ 
            \frac{1}{2} \| \vx^\T (\mI - \eta\mC)^n (\vw^\star - \vw_0) \|^2  
        \big].
\end{align}
With the near-zero initialization of the parameter of student and letting $\vw^{\star} = (\mX\mX^\T)^{-1}\mX\vt = N \mC^{-1}\mX\vt$, where $\vt$ is the target used in the pre-training of the teacher model, the loss is defined as:
\begin{align}
\label{eq:kd}
    \mathcal{L}^{(n)} = \E_{\vx \sim \mathcal{D}} \big[ 
            \frac{N}{2} \| \vx^\T (\mI - \eta\mC)^n \mC^{-1}\mX\vt \|^2  
        \big].
\end{align}
Here, $(\mI - \eta\mC)^n\mC^{-1} = \tilde{\mC}^{-1}$ is decomposed by SVD where the $i$-th eigenvalue is defined as:
\begin{align}
    \tilde{\lambda}_i = (1 - \eta \lambda_i)^n \lambda_i^{-1}.
\end{align}
With the appropriate $n$ using early stopping, the eigenvalue of the precision matrix $\mC^{-1}$ is filtered out for large eigenvalues while preserving small eigenvalues. 
Notice that this interpretation coincides with the notion of \thm~\ref{thm:low-rank}. The early stopping using a validation split may help to minimize the perturbation from small eigenvalues; however, it does not directly relate to anomaly detection.

Now, \eqn~\ref{eq:kd} can be seen as the difference between the squared Mahalanobis distance of $\vx$ and the squared Mahalanobis distance between a sample $\vx$ and a fixed point $\mX\vt$ using the filtered precision matrix $\tilde{\mC}^{-1}$, using the equation as follows:
\begin{align}
    \vx^\T \tilde{\mC}^{-1} \mX\vt 
    = \frac{1}{2} \vx^\T \tilde{\mC}^{-1} \vx - \frac{1}{2} (\vx - \mX\vt)^\T \tilde{\mC}^{-1} (\vx - \mX\vt) + \mathbf{c}
\end{align}
where the constant $\mathbf{c}$ is $\frac{1}{2} (\mX\vt)^\T \tilde{\mC}^{-1} (\mX\vt)$.

If $n \rightarrow \infty$, the loss is quickly conversed to zero with a learning rate of sufficiently small $\eta$.

\newpage
\section{Comparison with the state-of-the-art}

\tbl~\ref{tbl:mvtecad_pro_category} compares with the previous state-of-the-art method~\cite{Defard2020} varying the backbone networks and the hyper-parameter $k$ for each category. \textit{R18} and \textit{WR50} stand for ResNet-18 and Wide ResNet-50-2, respectively. The tailing number after a dash indicates $k$. Our method outperforms the competitive method in the majority of categories or shows competitive performances for some categories.

\begin{table*}[h!]
  \centering
  \caption{Comparison with the state-of-the-art for the anomaly segmentation task of the MVTec AD dataset using the PRO. Please see the text for details.}
  \label{tbl:mvtecad_pro_category}
  \begin{tabular}{lccccc}
  \toprule
  Model & \multicolumn{2}{c}{PaDiM~\cite{Defard2020}} & \multicolumn{3}{c}{Ours} \\
  \cmidrule(lr){1-1} \cmidrule(lr){2-3} \cmidrule(lr){4-6}
  Category   & R18-100 & WR50-550 & R18-100 & WR50-100 & WR50-300 \\ \midrule
  Carpet     & .960   &   .962   &   .970   &   .973   &\bf{.974}  \\
  Grid       & .909   &\bf{.946} &   .898   &   .908   &   .941    \\
  Leather    & .979   &   .978   &   .985   &   .985   &\bf{.987}  \\
  Tile       & .816   &\bf{.860} &   .788   &   .850   &   .859    \\
  Wood       & .903   &\bf{.911} &   .896   &   .908   &   .906    \\ \midrule
  Bottle     & .939   &   .948   &   .956   &   .961   &\bf{.962}  \\
  Cable      & .862   &   .888   &   .897   &   .896   &\bf{.915}  \\
  Capsule    & .919   &   .935   &   .945   &   .946   &\bf{.952}  \\
  Hazelnut   & .914   &   .926   &   .966   &   .966   &\bf{.970}  \\
  Metal nut  & .819   &   .856   &   .913   &\bf{.930} &\bf{.930}  \\
  Pill       & .906   &   .927   &   .916   &   .925   &\bf{.936}  \\
  Screw      & .913   &   .944   &   .940   &   .928   &\bf{.953}  \\
  Toothbrush & .923   &   .931   &\bf{.958} &   .953   &   .957    \\
  Transistor & .802   &   .845   &   .907   &   .924   &\bf{.929}  \\
  Zipper     & .947   &   .959   &   .957   &   .956   &\bf{.960}  \\ 
  \midrule
  Mean       & .901   &   .921   &   .926   &   .934   &\bf{.942}  \\
  \bottomrule
  \end{tabular}
\end{table*}

\section{Visualization}

\begin{figure}[t!]
    \begin{center}
        \includegraphics[width=\textwidth]{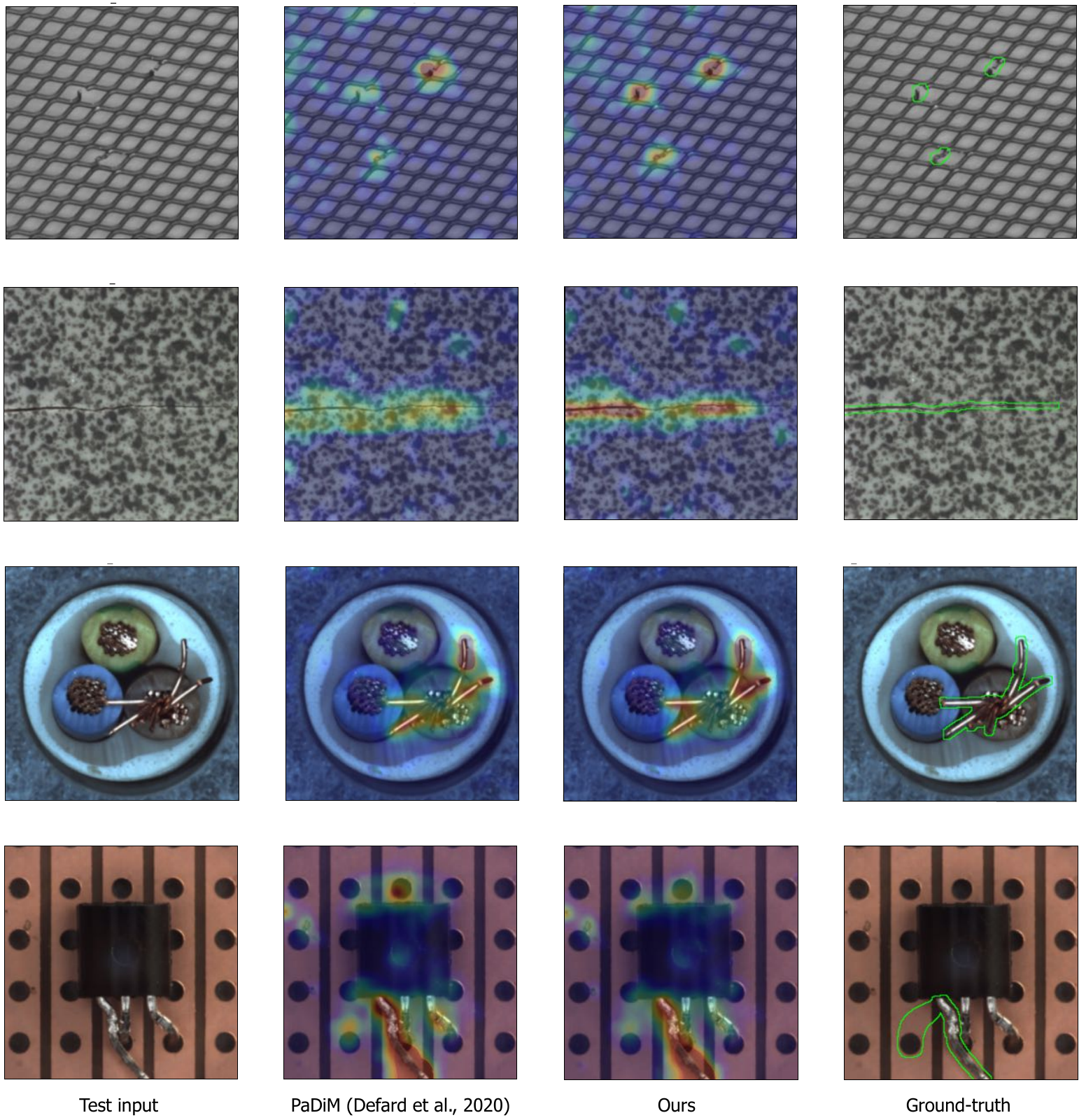}
    \end{center}
    \caption{The visualization using the jet color map is shown where the anomaly score is clamped in [0, 10]. The green lines in the fourth image indicate the ground-truth regions. From the top row, \textit{Grid}, \textit{Tile}, \textit{Cable}, and \textit{Transistor} are shown to compare with the state-of-the-art. Our method shows better detection of small or narrow regions, or reduces false-positive regions for these examples.}
    \label{fig:mvtecad_vs_visual}
\end{figure}

\begin{figure*}[t!]
    \begin{center}
        \includegraphics[width=\textwidth]{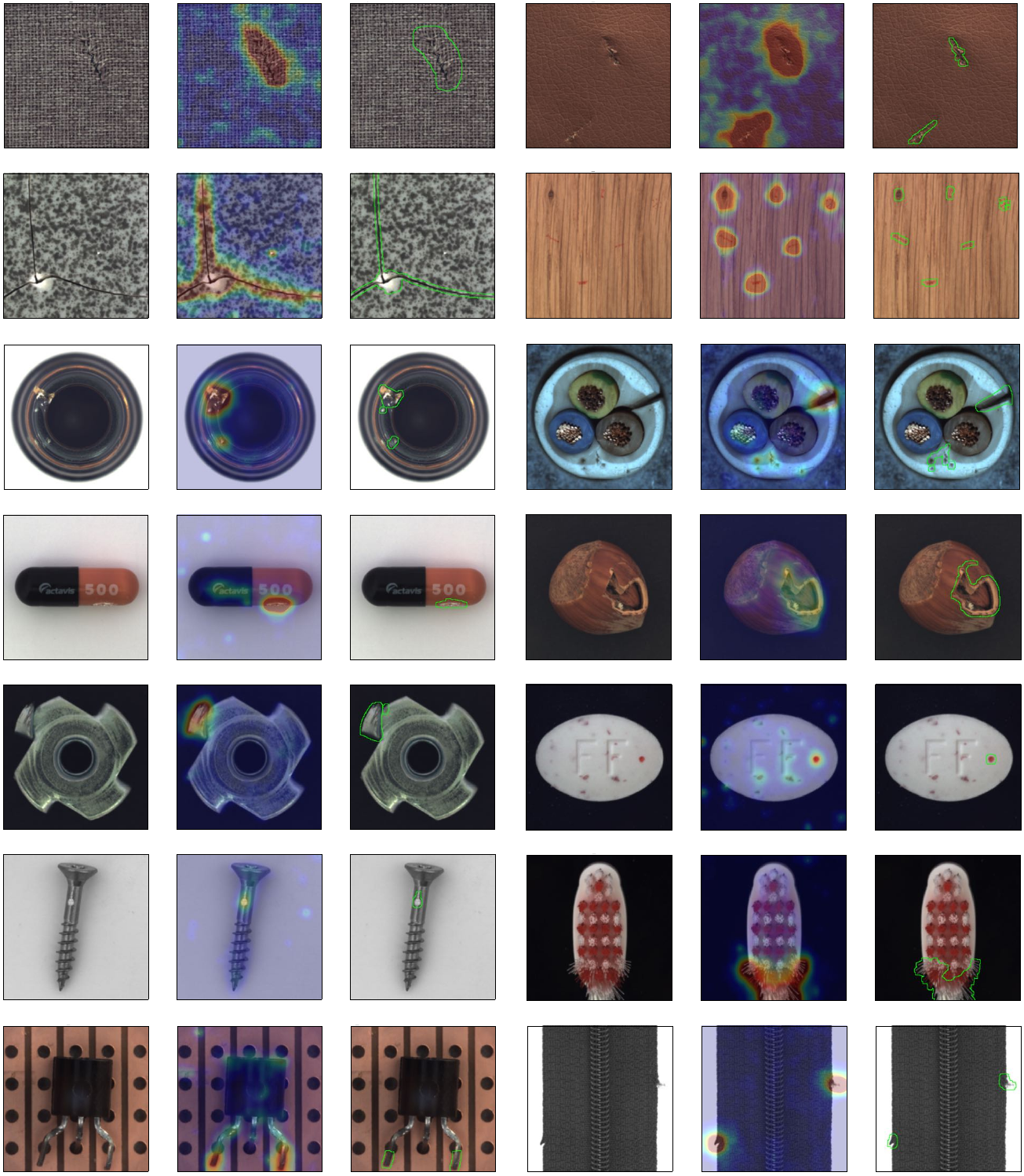}
    \end{center}
    \caption{The visualization of anomaly predictions for the MVTec AD. The first and fourth columns are test samples, the second and fifth columns are the anomaly prediction from our method using ResNet-18 and the $k$ of 100, and the third and sixth are the test samples with the boundaries (green solid lines) of ground-truth regions. The anomaly scores are clamped in [0, 10] and visualized using the jet color map (blue-yellow-red).}
    \label{fig:mvtecad_all_visual}
\end{figure*}

\fig~\ref{fig:mvtecad_vs_visual} shows extended examples from \fig~\ref{fig:mvtecad_grid_visual}, where the second and third columns are the anomaly prediction from PaDiM~\cite{Defard2020} and our method using ResNet-18 and the $k$ of 100, respectively.
\fig~\ref{fig:mvtecad_all_visual} show the visualization of the fourteen categories of the MVTec AD~\cite{Bergmann2019} except \textit{Grid}, which is previously shown in \fig~\ref{fig:mvtecad_grid_visual}.
The details can be referred in the caption of \fig~\ref{fig:mvtecad_all_visual}.

\end{document}